\documentclass{article}
\usepackage{ijcai26}

\usepackage{times}
\usepackage{soul}
\usepackage{url}
\usepackage[hidelinks]{hyperref}
\usepackage[utf8]{inputenc}
\usepackage[small]{caption}
\usepackage{graphicx}
\usepackage{amsmath}
\usepackage{amsthm}
\usepackage{booktabs}
\usepackage{algorithm}
\usepackage{algorithmic}
\usepackage[switch]{lineno}
\usepackage{xcolor}
\usepackage{multirow}
\usepackage{amssymb}

\title{FSX: Message Flow Sensitivity Enhanced Structural Explainer for Graph Neural Networks}

\author{
Bizu Feng$^1$
\and
Zhimu Yang$^2$\and
Shaode Yu$^{2*}$\And
Zixin Hu$^{1*}$\\
\affiliations
$^1$Fudan University\\
$^2$Communication University of China\\
\emails
bzfeng25@m.fudan.edu.cn, 
muyuzhierchengse@gmail.com, 
yushaodecuc@cuc.edu.cn, 
huzixin@fudan.edu.cn 
}

\begin{document}

\maketitle

\begin{abstract}
Despite the widespread success of Graph Neural Networks (GNNs), understanding the reasons behind their specific predictions remains challenging. Existing explainability methods face a trade-off that gradient‑based approaches are computationally efficient but often ignore structural interactions, while game‑theoretic techniques capture interactions at the cost of high computational overhead and potential deviation from the model’s true reasoning path. To address this gap, we propose FSX (Message Flow Sensitivity Enhanced Structural Explainer), a novel hybrid framework that synergistically combines the internal message flows of the model with a cooperative game approach applied to the external graph data. FSX first identifies critical message flows via a novel flow‑sensitivity analysis: during a single forward pass, it simulates localized node perturbations and measures the resulting changes in message flow intensities. These sensitivity‑ranked flows are then projected onto the input graph to define compact, semantically meaningful subgraphs. Within each subgraph, a flow‑aware cooperative game is conducted, where node contributions are evaluated fairly through a Shapley‑like value that incorporates both node‑feature importance and their roles in sustaining or destabilizing the identified critical flows. Extensive evaluation across multiple datasets and GNN architectures demonstrates that FSX achieves superior explanation fidelity with significantly reduced runtime, while providing unprecedented insights into the structural logic underlying model predictions—specifically, how important sub‑structures exert influence by governing the stability of key internal computational pathways.
\end{abstract}

\section{Introduction}

Graph neural networks (GNNs) have demonstrated strong performance on various graph tasks, enabling applications in domains such as social network recommendation \cite{fan2019graph,wu2022graph} and drug discovery \cite{jiang2021could}. This progress is driven by advances in key techniques, including graph convolutional networks \cite{kipf2016semi}, graph attention networks \cite{velivckovic2017graph}, and graph pooling methods \cite{lee2019self}. However, despite their effectiveness, these models are predominantly black-box in nature. Their internal reasoning processes are not transparent, and predictions are made without accompanying explanations. The lack of interpretability hinders a deeper understanding of model behavior and undermines trust, which prevents the deployment of GNNs in safety-critical applications. Therefore, explaining the predictions of GNNs has become an important research problem. 

In recent years, considerable research has been devoted to developing explanation methods for deep learning models, particularly in vision and language domains. However, the problem of explaining Graph Neural Networks (GNNs) is inherently more complex and less studied. A number of GNN-specific explainers have emerged from various perspectives, such as gradient-based methods~\cite{pope2019explainability}, perturbation-based methods~\cite{ying2019gnnexplainer,luo2020parameterized}, game-theoretic frameworks~\cite{10.1007/978-3-030-86520-7_19}, message-flow approaches~\cite{gui2023flowx}, and subgraph analysis techniques~\cite{yuan2021explainability}, among others. They often exhibit a critical limitation: they either prioritize computational efficiency by analyzing isolated components of the input graph, thereby overlooking the essential interactions within the model's internal reasoning process; or they attempt to attribute contributions via global game-theoretic constructs but suffer from prohibitive computational cost; alternatively, they may try to uncover internal interactions via local message flows, yet neglect possible explicit topological structures in the input graph and are severely affected by the model's own performance and external interference. We observe that truly faithful and efficient explanations can be achieved by directly leveraging the model's internal computational dynamics, especially message flows, to guide the identification and valuation of externally important sub-structures.

In this work, we propose FSX, a novel graph neural network explanation framework that identifies key subgraphs by directly linking the model's internal message flows to the input graph structure. Specifically, we introduce a flow-sensitivity analysis technique that performs local intervention on individual message vectors within the aggregation function. For a single message vector, this method simulates local perturbations in a forward pass to rank the sensitivity of message flows, effectively avoiding instability caused by training influences. Since the stability of these critical flows is governed by specific sub-structures in the input graph, we propose mapping high-sensitivity flows onto the graph to identify candidate explanatory subgraphs. To fairly quantify the contribution of each node within a subgraph, we formulate a flow-aware cooperative game model. This computes a Shapley-like value that integrates both node feature importance and its role in maintaining critical flows, rather than using a Shapley value that treats flows as players in a graph cooperative game. Furthermore, we design an efficient approximation scheme that confines game-theoretic evaluation to the compact subgraphs derived from flow analysis, rather than the entire graph, significantly reducing computational complexity. Overall, this study presents the first hybrid approach that cohesively integrates internal model dynamics with external graph structure for GNN explanation. Extensive experiments on multiple benchmark datasets and GNN architectures show that FSX not only generates explanations with comparable performance to existing interpreters, but also significantly reduces computational time, effectively bridging the gap between explanation fidelity and efficiency.

\section{Related work}
\subsection{Graph Neural Networks}

Graph neural networks have become a foundational tool for learning on graph-structured data, achieving state-of-the-art results across numerous tasks. A series of architectures have been proposed to learn effective node and graph representations, notable examples include Graph Convolutional Networks (GCNs)~\cite{kipf2016semi}, which perform efficient spectral convolutions through localized first-order approximations; Graph Attention Networks (GATs)~\cite{velivckovic2017graph}, which utilize attention mechanisms to assign varying importance to neighbors; and Graph Isomorphism Networks (GINs)~\cite{xu2018how}, which offer theoretically maximal expressive power within the message-passing framework. These models predominantly operate under a message-passing paradigm, where each node iteratively updates its representation by aggregating (``passing'') messages from its immediate neighbors. To concretize this process, we take the Graph Isomorphism Network (GIN)~\cite{xu2018how} as a representative example. Formally, let an undirected graph be represented by an adjacency matrix $\mathbf{A} \in \{0,1\}^{N \times N}$ and a node feature matrix $\mathbf{X} \in \mathbb{R}^{N \times F}$, where $N$ is the number of nodes and $F$ is the feature dimension. A single layer of GIN performs the following aggregation operation:
\begin{equation}
\label{eq:gin_layer}
\mathbf{h}_i^{(l+1)} = \text{MLP}^{(l)}\left( (1 + \epsilon^{(l)}) \cdot \mathbf{h}_i^{(l)} + \sum_{j \in \mathcal{N}(i)} \mathbf{h}_j^{(l)} \right).
\end{equation}
Here, \(\mathbf{h}_i^{(l)}\) is the representation of node \(i\) at layer \(l\), with \(\mathbf{h}_i^{(0)} = \mathbf{x}_i\). \(\mathcal{N}(i)\) denotes the set of neighboring nodes of node \(i\). The learnable parameter \(\epsilon^{(l)}\) is a scalar that controls the relative importance of the node's own features versus its neighbors' aggregated features. \(\text{MLP}^{(l)}(\cdot)\) denotes a multi-layer perceptron at layer \(l\), which provides the non-linearity and transformation. This aggregation scheme, particularly the use of an injective function like the MLP on a sum aggregation, is theoretically grounded to achieve maximum discriminative power (expressive power) among all message-passing GNNs for distinguishing graph structures.

\subsection{Explainability in Graph Neural Networks}

The ability to explain GNN predictions is critical for fostering trust and ensuring responsible deployment. Existing post-hoc explanation methods for GNNs can be broadly categorized based on their reliance on the GNN’s internal workings, largely falling into two distinct paradigms.

The first, model-agnostic (or black-box) paradigm, treats the GNN as an opaque function. These methods explain predictions by probing the model with perturbations to the input graph and observing changes in the output. A prominent family of methods within this category is perturbation-based explainers. For instance, GNNExplainer ~\cite{ying2019gnnexplainer} learns a soft mask over edges and node features to maximize mutual information with the original prediction. GraphEXT~\cite{wu2025explainable} is a theoretically novel variant within this paradigm. It introduces the economic concept of externality into graph structure modeling and quantifies node importance via the Shapley value under externalities from cooperative game theory. However, GraphEXT operates entirely from an external perturbation and game-theoretic perspective, failing to leverage any of the GNN's internal computational mechanisms, such as message flows. This disconnects its explanations from the model's actual reasoning process. Other notable directions include counterfactual explanation methods (e.g., CLEAR~\cite{ma2022clear}), which seek minimal changes to the input that would alter the prediction, and surrogate model-based approaches ). While efficient and intuitive, these model-agnostic methods inherently lack access to the internal reasoning process of the GNN.

In contrast, the second, model-internal paradigm, seeks to open the black box by directly analyzing the GNN’s internal computational mechanisms, such as gradients, attention weights, or message flows. More recent advances focus explicitly on the message-passing mechanism. For example, FlowX~\cite{gui2023flowx} pioneered the analysis of message flows, treating them as collaborative players in a cooperative game and approximating their importance via Shapley values. However, FlowX faces two primary limitations: (1) it models flows in isolation, without directly linking them to the structural subgraphs in the input that govern their stability; (2) its computational cost remains high, as it requires sampling over all possible layer-edge permutations to approximate flow-level Shapley values. Other recent works, such as Faithful and Accurate Self-Attention Attribution~\cite{Shin_Li_Cao_Shin_2025} and GOAt~\cite{lu2024goat}, develop principled attribution techniques for attention-based GNNs. These methods offer the promise of higher fidelity by design, as they are grounded in the model’s actual operations, but they often face significant challenges in terms of computational cost or structural interpretation.

Recognizing the complementary strengths and weaknesses of these two paradigms, a few hybrid approaches have begun to emerge, seeking a preliminary integration. Some methods use internal signals (like gradients) to guide or initialize a subsequent perturbation-based search. Others employ information from multiple GNN layers to constrain the search space for important subgraphs. However, these integrations often remain shallow. In particular, existing works have failed to deeply integrate internal flow analysis, as in FlowX, with game-theoretic evaluation based on structural interactions, as in GraphEXT, into a unified framework. Consequently, a deep, synergistic fusion that leverages internal message flows to directly and efficiently guide the identification and valuation of external subgraph structures is still lacking. This gap motivates our work.

\section{Method}
\begin{figure*}
    \centering
    \includegraphics[width=1\linewidth]{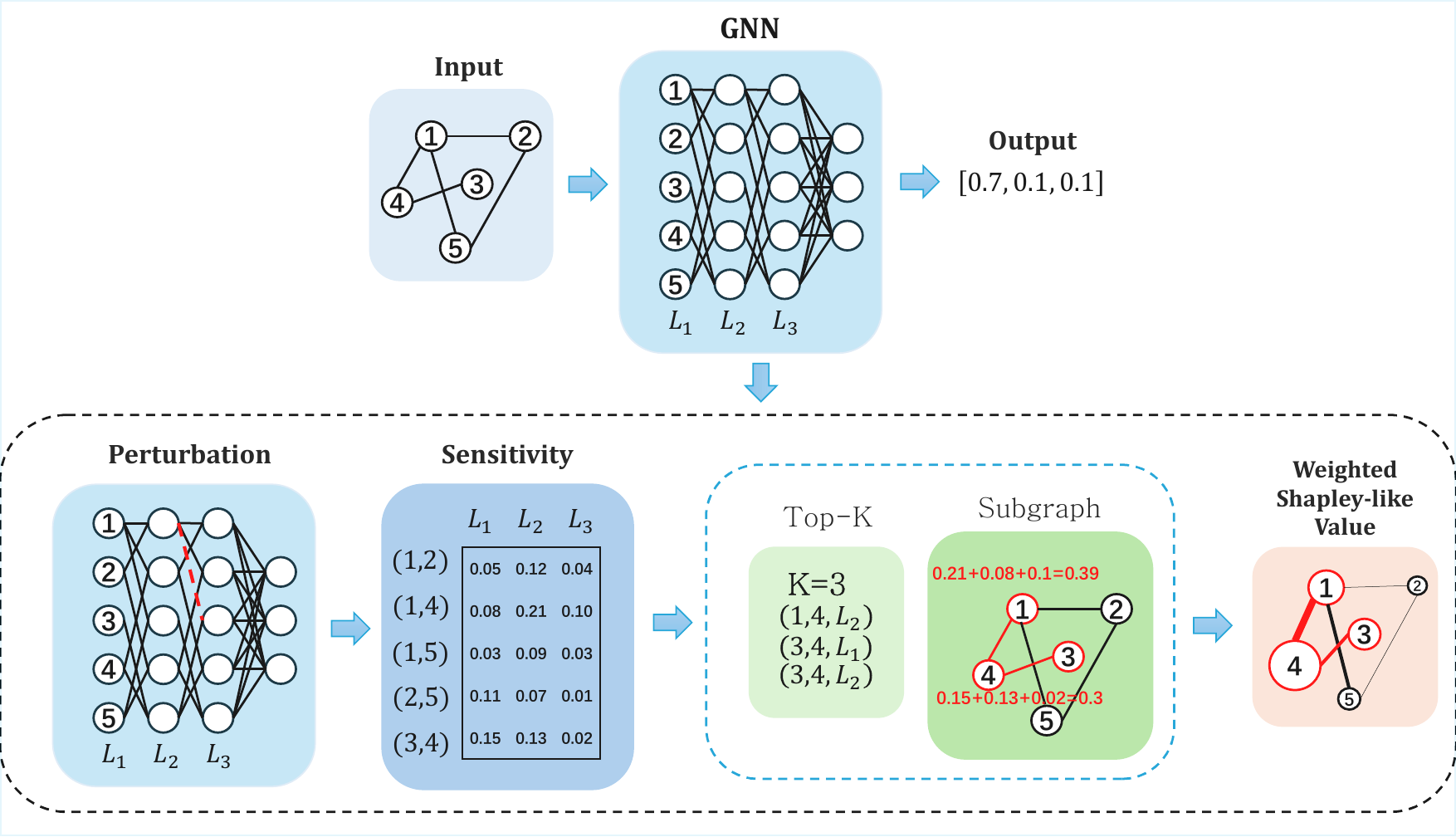}
    \caption{An illustration of FSX}
    \label{fig:FSX}
\end{figure*}

Existing GNN explainers predominantly fall into two categories: model-agnostic methods that treat the GNN as a black box and model-internal methods that analyze its computational mechanisms. However, the former often lacks fidelity to the model's true reasoning, while the latter suffers from high computational cost. To bridge this gap, we argue that a faithful and efficient explanation must synergistically combine the internal computational dynamics of the GNN with the external graph structure. We propose FSX (Message Flow Sensitivity Enhanced Structural Explainer), a novel hybrid framework that achieves this by using the model's internal message flows to directly guide the identification and valuation of important subgraphs. Specifically, FSX operates in two cohesive stages: (1) It first performs a flow-sensitivity analysis via localized perturbations within a forward pass to identify critical message flows governing the prediction. (2) These critical flows are then mapped to the input graph to define compact candidate subgraphs, within which a flow-aware cooperative game is conducted to fairly evaluate node contributions. By tightly coupling internal flow dynamics with external structural evaluation, FSX efficiently generates high-fidelity subgraph-level explanations.The workflow of the proposed method is illustrated in Figure \ref{fig:FSX}.
\subsection{Flow-Sensitivity Analysis via Localized Perturbation}
The inaugural stage of FSX is designed to pierce the opaque reasoning of GNNs by directly interrogating their internal computational pathways. We posit that the causal drivers of a GNN's prediction are best understood by intervening directly on the fundamental units of its operation: the message flows that propagate information between nodes across layers. This approach shifts the explanatory paradigm from correlation to causal intervention.

Formally, consider an $L$-layer GNN model $f$ trained for graph classification. Given an input graph $\mathcal{G} = (\mathcal{V}, \mathcal{E})$ with node feature matrix $\mathbf{X}$ and adjacency matrix $\mathbf{A}$, let $s_{\text{orig}} = f(\mathcal{G})[c_{\text{target}}]$ denote the model's original output logit for a target class $c_{\text{target}}$. Our objective is to compute a sensitivity score for every message flow triplet $(u, v, l)$, where $(u, v) \in \mathcal{E}$ and $l \in \{1, \ldots, L\}$.

The core of our analysis is a single, engineered forward pass of the GNN model. For each candidate triplet $(u, v, l)$, we simulate a localized and targeted perturbation. Within the message aggregation function of the $l$-th layer, we intercept the message $\mathbf{h}_u^{(l-1)}$ destined from $u$ to $v$ and apply a weakening perturbation, transforming it to $\tilde{\mathbf{h}}_u^{(l-1)} = \gamma \cdot \mathbf{h}_u^{(l-1)}$, where $\gamma$ is a damping factor ($0\leq \gamma < 1$). This perturbation is exclusively local: it alters only the message on the edge $(u,v)$ at layer $l$; all other messages remain unchanged. This precision is achieved by embedding conditional logic into the GNN's message-passing kernel.

Following this localized intervention, the perturbed model completes its forward propagation, yielding a perturbed prediction score $s_{\text{pert}}^{(u,v,l)}$. The flow sensitivity $S(u,v,l)$ is then defined as:
\begin{equation}
\label{eq:sensitivity}
S(u,v,l) = \left| s_{\text{orig}} - s_{\text{pert}}^{(u,v,l)} \right|.
\end{equation}
This metric captures the magnitude of the prediction change attributable to the weakening of the specific message flow $(u,v,l)$. A large value of $S(u,v,l)$ signifies that the information carried along that path is a critical determinant of the model's prediction; a sensitivity score near zero implies the message flow is redundant or inconsequential.

\begin{algorithm}[t]
\caption{Flow-Sensitivity Analysis via Localized Perturbation}
\label{alg:flow_sensitivity}
\textbf{Input:} Trained $L$-layer GNN $f$, input graph $\mathcal{G}=(\mathcal{V}, \mathcal{E})$, node features $\mathbf{X}$, target class $c_{\text{target}}$, damping factor $\gamma$;\\
\textbf{Output:} Sensitivity map $\mathcal{S}$;
\begin{algorithmic}[1]
\STATE $s_{\text{orig}} \gets f(\mathcal{G})[c_{\text{target}}]$
\STATE $\mathcal{S} \gets \varnothing$
\FOR{$l = 1$ \textbf{to} $L$}
    \FOR{$(u, v) \in \mathcal{E}$}
        \STATE Create $f_{\text{tmp}}$ with modified layer $l$:
        \STATE \hspace{\algorithmicindent} Replace $\mathbf{h}_u^{(l-1)}$ with $\gamma \cdot \mathbf{h}_u^{(l-1)}$ in aggregation at node $v$
        \STATE $s_{\text{pert}} \gets f_{\text{tmp}}(\mathcal{G})[c_{\text{target}}]$
        \STATE $S(u,v,l) \gets | s_{\text{orig}} - s_{\text{pert}} |$
        \STATE $\mathcal{S} \gets \mathcal{S} \cup \{(u, v, l, S(u,v,l))\}$
    \ENDFOR
\ENDFOR
\RETURN $\mathcal{S}$
\end{algorithmic}
\end{algorithm}

By systematically executing this procedure we construct a comprehensive message flow triplet $\mathcal{S} = \{ S(u,v,l) \}$. This  provides a fine-grained, quantitative ranking of all internal message flows based on their estimated causal contribution to the GNN's decision for the given input graph. The map $\mathcal{S}$ serves as the pivotal, information-rich interface between the model's internal dynamics and the subsequent stage of our framework, directly guiding the identification of critical substructures within the input graph.

\subsection{Flow-Aware Subgraph Identification and Valuation}
Building upon the fine-grained flow sensitivity map $\mathcal{S}$ obtained from Sec. 3.1, FSX's second stage accomplishes a crucial translation: it transforms the internal, causal view of model computation into an externally interpretable, graph-level explanation endowed with quantitatively validated node contributions. This stage is architected to bridge the gap between the pinpoint precision of flow-level analysis and the human-intelligible need for structural explanations. It operates through three sequentially dependent and logically cohesive components: (1) the distillation of a compact, semantically rich key subgraph directly from the most sensitive message flows, (2) the formulation of a novel flow-aware cooperative game on this subgraph reflect the stability of the identified critical information pathways, and (3) the efficient approximation of a weighted Shapley-like value that fairly allocates explanatory power to nodes within this causal framework.

\subsubsection{From Flows to Subgraphs: Distilling Causal Pathways}
The sensitivity map $\mathcal{S}$ provides a ranking of every message flow triplet $(u,v,l)$ by its estimated causal impact. The first step is to filter this extensive data to isolate the most consequential information channels. Let $K$ be a predefined budget, representing the desired explanation complexity. We select the set $\mathcal{T}_{\text{key}}$ of top-$K$ triples:
\[
\mathcal{T}_{\text{key}} = \operatorname{TopK}_{(u,v,l) \in \mathcal{S}} \; S(u,v,l).
\]
This selection captures the message flows whose perturbation caused the largest deviation in the model's prediction, implying they lie on the GNN's primary reasoning paths for the specific input.

The key subgraph $G_{\text{key}} = (V_{\text{key}}, E_{\text{key}})$ is then \textit{induced} from these pivotal triples:
\[
V_{\text{key}} = \bigcup_{(u,v,l) \in \mathcal{T}_{\text{key}}} \{u, v\}, \quad E_{\text{key}} = \bigcup_{(u,v,l) \in \mathcal{T}_{\text{key}}} \{(u, v)\}.
\]
This inductive construction guarantees that $G_{\text{key}}$ contains precisely the nodes and edges that physically instantiate the most causally significant information flows within the GNN's computational graph.

\begin{algorithm}[t]
\caption{Weighted Shapley-like Value Approximation for FSX}
\label{alg:weighted_shapley}
\textbf{Input:} Key subgraph $G_{key} = (V_{key}, E_{key})$, value function $\nu$, flow weights $w$, sampling iterations $M$, weighting function $g(\cdot)$;\\
\textbf{Output:} Node contribution scores $\hat{\phi}_i$ for all $i \in V_{key}$;

\begin{algorithmic}[1]
\STATE Initialize $\phi_i \gets 0$, $W_i \gets 0$ for all $i \in V_{key}$
\FOR{$iter = 1$ to $M$}
    \STATE Generate a random permutation $\pi$ of $V_{key}$.
    \STATE Initialize $S_{\text{prev}} \gets \emptyset$.
    \FOR{each node $i$ in the order of permutation $\pi$}
        \STATE $S_{\text{curr}} \gets S_{\text{prev}} \cup \{i\}$.
        \STATE Compute marginal contribution: 
        \STATE $mc \gets \nu(S_{\text{curr}}) - \nu(S_{\text{prev}})$.
        \STATE Compute coalition flow weight: 
        \STATE $I_{\text{curr}} \gets \sum_{(u,v) \in E(S_{\text{curr}})} w(u,v)$.
        \STATE Compute weight: $\beta \gets g(I_{\text{curr}})$.
        \STATE $\phi_i \gets \phi_i + \beta \cdot mc$.
        \STATE $W_i \gets W_i + \beta$.
        \STATE $S_{\text{prev}} \gets S_{\text{curr}}$.
    \ENDFOR
\ENDFOR
\FOR{each node $i \in V_{key}$}
    \IF{$W_i > 0$}
        \STATE $\hat{\phi}_i \gets \phi_i / W_i$.
    \ELSE
        \STATE $\hat{\phi}_i \gets 0$.
    \ENDIF
\ENDFOR
\RETURN $\{\hat{\phi}_i \mid i \in V_{key}\}$.
\end{algorithmic}
\end{algorithm}

\subsubsection{Flow-Aware Cooperative Game Formulation: Valuing Structural Stability}
To quantify the contribution of each node within $G_{\text{key}}$, we model the explanation task as a cooperative game $(V_{\text{key}}, \nu)$. The key innovation lies in the design of the value function $\nu(S)$, which must reflect not only a coalition $S \subseteq V_{\text{key}}$'s ability to produce a similar prediction but, more fundamentally, its capacity to sustain the critical internal message flows identified in Stage 1.

We first derive an information flow weight $w(u,v)$ for each edge $(u,v) \in E_{\text{key}}$. This weight synthesizes the sensitivity of all selected flows that traverse this edge across different layers:
\begin{equation}
w(u,v) = \sum_{l \in L_{\text{key}}(u,v)} S(u,v,l).
\label{eq:edge_weight}
\end{equation}
where $L_{\text{key}}(u,v) = \{ l \mid (u,v,l) \in \mathcal{T}_{\text{key}} \} $.Here, $w(u,v)$ aggregates the causal importance of the edge $(u,v)$ acting as a message carrier at various depths of the network. For edges not in $E_{\text{key}}$, we set $w(u,v)=0$, effectively ignoring their contribution in this localized analysis. For any node coalition $S$, we define its coalition flow weight $I(S)$ as the sum of flow weights over all edges in the subgraph induced by $S$:
\[
I(S) = \sum_{(u,v) \in E(S)} w(u,v),
\]
where $E(S)$ denotes the edge set of the subgraph induced by $S$. $I(S)$ serves as a succinct measure of how much of the total critical flow mass is encompassed within coalition $S$.

The value function $\nu(S)$ is then a composition of two factors:
\[
\nu(S) = g\left( I(S) \right) \cdot f\left(G[S]\right)[c_{\text{target}}].
\]
The term $f\left(G[S]\right)[c_{\text{target}}]$ is the GNN's prediction score when the input is restricted to the subgraph induced by $S$ (non-member nodes and edges are masked out), capturing the coalition's direct predictive power. The modulating function $g(\cdot)$ is a non-negative, monotonically increasing weighting function of $I(S)$. We instantiate $g(\cdot)$ as a sigmoid function:
\[
g(x) = \frac{1}{1 + \exp(-(x - x_0)/\tau)}.
\]
Here, $x_0$ is a centrality parameter (typically set to the median of observed $I(S)$ values) and $\tau > 0$ is a temperature controlling the sharpness of the transition. This design embodies a core principle: a coalition's value is \textit{amplified} if it contains a high concentration of edges that are vital to the model's internal message passing. The sigmoid function implements a soft, differentiable threshold, ensuring that coalitions sustaining a sufficient mass of critical flows receive proportionally greater consideration in the contribution allocation, thereby directly incorporating the causal flow information into the game-theoretic framework.

\subsubsection{Efficient Weighted Shapley-like Value Approximation: Bias for Causal Fidelity}
The final step is to fairly distribute the total explanatory value among individual nodes in $V_{\text{key}}$ based on their marginal contributions within the flow-aware game. The classical Shapley value provides an axiomatic solution but assumes all player orderings are equally likely. This assumption is at odds with our objective: we want to emphasize contributions made in contexts (i.e., node orderings/coalitions) where the critical flows are active.

Therefore, we define the importance of node $i$ via a Weighted Shapley-like Value $\phi^{\text{IFW}}_i$:
\begin{equation}
\phi^{\text{IFW}}_i = \frac{ \sum_{\pi \in \Pi} \; g\!\left(I(P_i^\pi \cup \{i\})\right) \cdot \left[ \nu(P_i^\pi \cup \{i\}) - \nu(P_i^\pi) \right] }{ \sum_{\pi \in \Pi} g\!\left(I(P_i^\pi \cup \{i\})\right) },
\label{eq:weighted_shapley}
\end{equation}
where $\Pi$ is a set of sampled permutations of $V_{\text{key}}$, $P_i^\pi$ denotes the set of players preceding $i$ in permutation $\pi$, and $g(I(P_i^\pi \cup \{i\}))$ is the same flow-weight-aware weighting function from the value definition.

This formulation introduces a profound shift: the marginal contribution of node $i$ when joining a coalition $P_i^\pi$ is weighted by the \textit{flow importance of the resulting coalition} $P_i^\pi \cup \{i\}$. This means contributions made in the context of forming or extending a coalition that is rich in critical flows are given higher credence. It effectively biases the explanation towards recognizing nodes that are instrumental in assembling or stabilizing the high-flow-weight sub-structures that drive the GNN's prediction, ensuring the final node-level explanation remains faithful to the internal causal narrative uncovered in Stage 1.

Computing Eq.~\eqref{eq:weighted_shapley} exactly requires averaging over all $|V_{\text{key}}|!$ permutations, which is intractable. The efficiency of FSX stems from two key designs: First, the game is played on the small key subgraph $G_{\text{key}}$, drastically reducing the number of players. Second, we employ a Monte Carlo approximation (detailed in Algorithm~\ref{alg:weighted_shapley}) that samples permutations and leverages the efficient evaluation of $\nu(S)$ (a single forward pass on a small subgraph). This makes the computation of $\phi^{\text{IFW}}_i$ for all $i \in V_{\text{key}}$ highly practical. The output is a set of node contribution scores $\{\phi^{\text{IFW}}_i\}$ that collectively form the final explanation: a quantitative assessment of each node's role within the causally salient subgraph $G_{\text{key}}$.

The algorithm runs in $O(M \cdot |V_{key}| \cdot T_f)$, where $T_f$ is the cost of evaluating $\nu$. Since $|V_{key}|$ is small (as $G_{key}$ is compact) and $\nu$ only requires a forward pass on a small subgraph, the computation is highly efficient. The output is a set of node contribution scores $\{\hat{\phi}_i\}$ that constitute the final explanation, highlighting nodes that are pivotal within the critical flow-informed substructure.

\section{Experiment}
We performed extensive experiments to evaluate the effectiveness and efficiency of our proposed FSX framework. We focused on assessing whether FSX provides more faithful explanations and is computationally efficient.
\begin{figure*}
    \centering
    \includegraphics[width=1\linewidth]{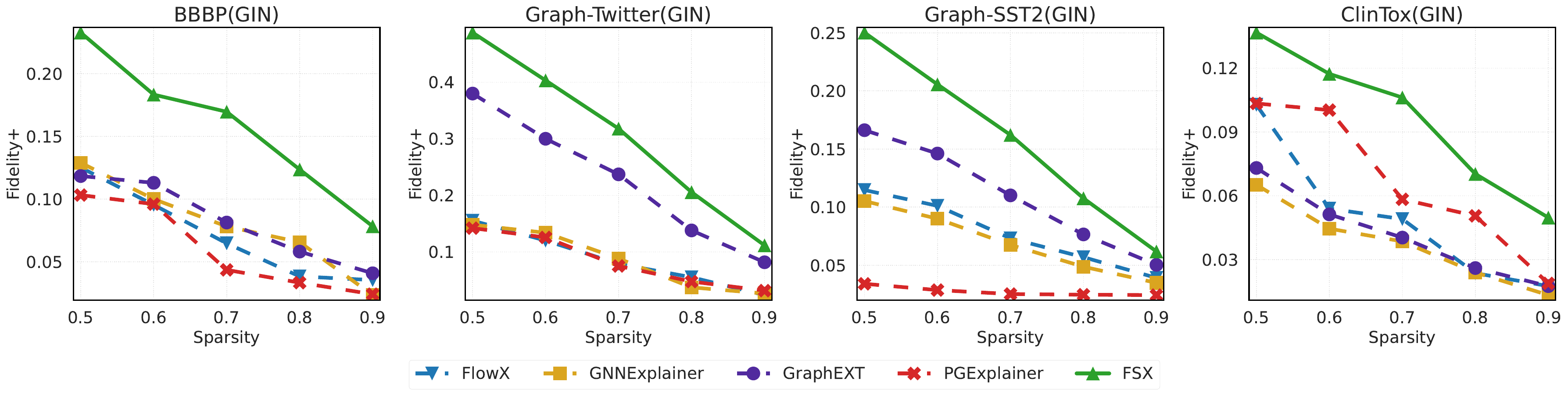}
    \caption{Fidelity+ values were evaluated on 4 datasets using GINs under different sparsity levels, where a higher Fidelity+ value indicates better performance.}
    \label{fig:fidelity+}
\end{figure*}

\subsection{Datasets and Baselines}
To comprehensively evaluate FSX, we employed four real-world benchmark datasets. These datasets cover diverse domains and graph classification tasks, allowing for a thorough assessment of the explainer's performance on various graph types. Subsequently, we chose several state-of-the-art methods for graph structure explanation as baselines, including FlowX\cite{gui2023flowx} , GNNExplainer\cite{ying2019gnnexplainer} , GraphEXT\cite{wu2025explainable} , and PGExplainer\cite{luo2020parameterized}. For all experiments, a 3-layer Graph Isomorphism Network (GIN) model was utilized as the base GNN. The datasets and baseline implementations were based on the DIG Library\cite{liu2021dig}. The detailed statistical properties of these datasets are summarized in Table 1.

BBBP (Blood-Brain Barrier Penetration) and ClinTox\cite{wu2018moleculenet} are molecular property prediction datasets from the Therapeutics Data Commons (TDC). Each graph in these datasets represents a molecule, with nodes as atoms and edges as chemical bonds. Graph labels for BBBP indicate whether a compound can penetrate the blood-brain barrier, while ClinTox labels denote toxicity and clinical trial outcomes. These datasets are crucial for evaluating explainer in the context of drug discovery and medicinal chemistry.

Graph-SST2 and Graph-Twitter\cite{yuan2022explainability} are graph-of-words datasets designed for sentiment analysis, where textual data is transformed into graph representations. In these graphs, nodes correspond to words, and edges represent their co-occurrence or syntactic relationships. Each graph is classified based on its sentiment polarity (positive or negative). These datasets allow us to assess explainer's ability to interpret explanations in natural language processing tasks.

\subsection{Settings and evaluation metrics}
For our FSX framework, the number of top-sensitive flows K for constructing the key subgraph is set as 20\% of the total possible flows. In the Shapley Algorithm, we utilize $M = 100$ Monte Carlo samples for both FSX and GraphEXT. For FlowX, the number of Monte Carlo samples is dynamically determined, with a maximum limit of 100. The framework is implemented in PyTorch 2.5.1 and PyTorch Geometric. All experiments are conducted on a single NVIDIA RTX 4090 (24GB) GPU. To evaluate the quality of explanations, we adopt three widely recognized metrics: Fidelity+ measures the average drop in predicted probability for the original class when the top-$k$ important nodes (according to the explanation) are removed (features masked), where a higher value is better. Fidelity- quantifies the average drop in predicted probability when only the top-$k$ important nodes are retained, with a lower value being preferable. Lastly, Sparsity represents the proportion of nodes selected as important, where lower sparsity indicates a more concise explanation.

\begin{figure*}
    \centering
    \includegraphics[width=1\linewidth]{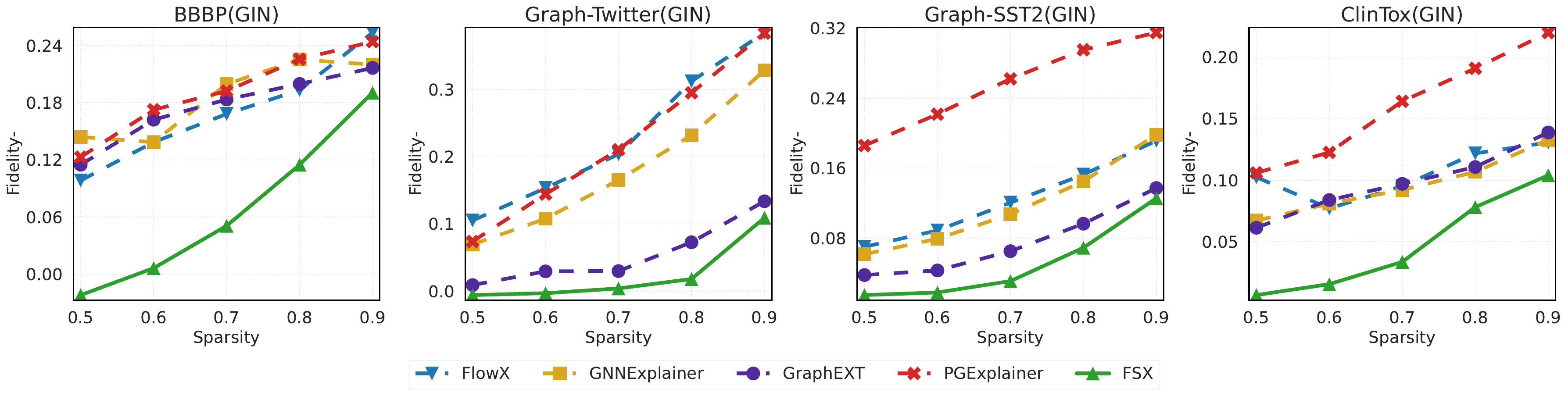}
    \caption{Fidelity- values were evaluated on 4 datasets using GINs under different sparsity levels, where a lower Fidelity- value indicates better performance.}
    \label{fig:fidelity-}
\end{figure*}

\subsection{Explanation Comparison}
We employ $\text{Fidelity}^+$, $\text{Fidelity}^-$, and $\text{Sparsity}$ as core evaluation metrics. Given a graph classification model  $f(\cdot)$, a sample graph $G = (X, A)$, the model's predicted class $y$, and the important node mask $M$ generated by an explanation method, the metrics are defined as:
\begin{equation}
    \text{Fidelity}^+ = \frac{1}{N} \sum_{i=1}^{N} \left[ f(G)_y - f(G \setminus M^{(i)})_y \right]
\end{equation}
\begin{equation}
    \text{Fidelity}^- = \frac{1}{N} \sum_{i=1}^{N} \left[ f(G)_y - f(G_{M^{(i)}})_y \right]
\end{equation}
\begin{equation}
    \text{Sparsity} = \frac{1}{N} \sum_{i=1}^{N} \left( 1 - \frac{ \| M^{(i)} \|_0 }{ |V^{(i)}| } \right)
\end{equation}
where $G \setminus M$ denotes the subgraph obtained by removing the nodes corresponding to mask $M$, $G_M$ represents the subgraph formed by retaining only the nodes indicated by the mask, $\| M \|_0$ is the number of important nodes, and $|V|$ is the total number of nodes.

We conduct experiments on several public graph datasets using a trained Graph Isomorphism Network (GIN) model. We first control the sparsity of explanations and compare the performance of different methods in terms of $\text{Fidelity}^+$. A high $\text{Fidelity}^+$ score indicates that the identified structures are necessary for the prediction. As shown in Figure~\ref{fig:fidelity+}, our proposed method consistently outperforms the baseline comparisons in $\text{Fidelity}^+$ under varying sparsity constraints. The advantage is particularly more pronounced in the high sparsity region ($\text{Sparsity} > 0.8$) on synthetic datasets.

Furthermore, we evaluate performance using the $\text{Fidelity}^-$ metric, which measures the sufficiency of the explanatory structures. A lower $\text{Fidelity}^-$ value suggests that the identified subgraphs are sufficient to support the original prediction. As shown in Figure~\ref{fig:fidelity-}, on most real-world datasets, our method achieves significantly lower $\text{Fidelity}^-$ than baseline methods at comparable sparsity levels.

Considering both metrics together, our method exhibits a better balance between $\text{Fidelity}^+$ and $\text{Fidelity}^-$, indicating it can both identify indispensable nodes and avoid omitting important decision evidence. Moreover, on graph datasets for sentiment classification, our method shows stronger explanatory consistency, which we attribute to its explicit modeling of synergistic contributions between node-local topology and features.

\begin{figure}
    \centering
    \includegraphics[width=1\linewidth]{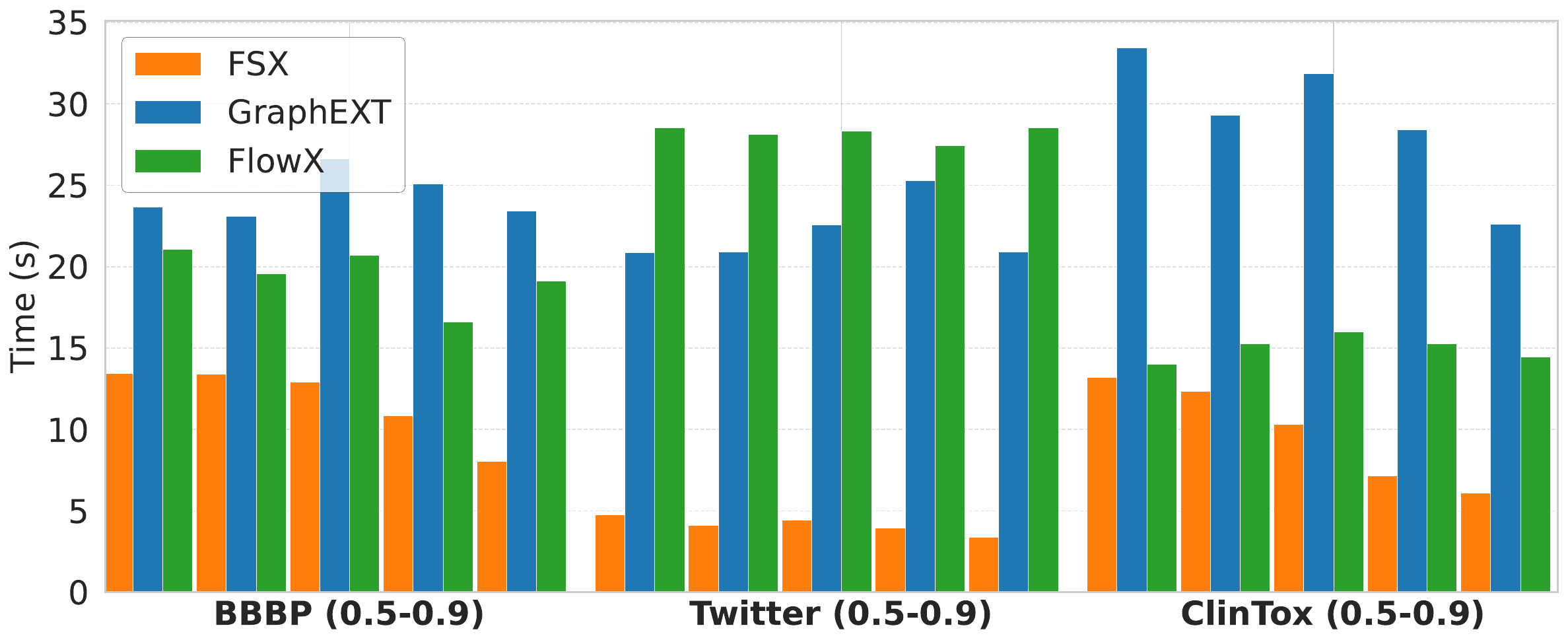}
    \caption{Average Time}
    \label{fig:time}
\end{figure}

Several key observations explain FSX’s superior performance. First, compared to purely perturbation-based methods like GNNExplainer and PGExplainer, FSX benefits from its internal flow-sensitivity analysis, which directly pinpoints causally important message flows. Second, compared to game-theoretic methods like  GraphEXT FSX gains significant efficiency from its two-stage process, avoiding the prohibitive cost of evaluating the Shapley value on the entire graph. Third, even compared to the flow-based explainer FlowX, FSX shows clear improvements by using flows to guide structural identification and employing a flow-weighted game to assess node contributions, resulting in explanations that are both faithful to the model’s internal dynamics and structurally coherent.

\subsection{Efficiency Analysis}
In addition to faithfulness and compactness, the computational efficiency of an explanation method is crucial for its practical applicability. To this end, we conduct a runtime analysis by measuring the average explanation time per sample across all test graphs. As illustrated in Figure~\ref{fig:time}, our proposed method requires substantially less time compared to the baseline approaches. This significant efficiency gain stems primarily from our strategy of leveraging information-flow sensitivity to rapidly locate a small, influential subgraph, thereby circumventing the high computational overhead associated with whole-graph processing. The result confirms that our method not only provides high-quality explanations but also maintains high computational efficiency, making it suitable for real-world and large-scale applications.

\section{Conclusion}
In this work, we address the critical challenge of explaining the predictions of Graph Neural Networks, which often operate as inscrutable black boxes. Moving beyond the limitations of existing model-agnostic or model-internal paradigms, we propose FSX, a novel hybrid explanation framework that synergistically integrates the GNN’s internal message flow dynamics with external graph structure valuation. This work establishes a systematic approach for generating faithful and efficient graph-level explanations, encompassing a flow-sensitivity analysis for causal insight, a flow-aware cooperative game for fair contribution assessment, and an efficient approximation scheme. Extensive experiments demonstrate that FSX not only achieves superior explanation fidelity compared to state-of-the-art baselines but does so with significantly reduced computational cost, effectively bridging the gap between interpretability and efficiency. We believe this work underscores the importance of tightly coupling a model’s internal reasoning mechanisms with the explanation process and hope it paves the way for more principled, efficient, and insightful explainable AI for graph-structured data.

\bibliographystyle{named}
\bibliography{ijcai26}

\end{document}